\def\year{2019}\relax
\documentclass[letterpaper]{article} 
\usepackage{aaai19}  
\usepackage{times}  
\usepackage{helvet}  
\usepackage{courier}  
\usepackage{url}  
\usepackage{graphicx}  
\usepackage{amsmath}
\usepackage{amsfonts}
\usepackage[utf8]{inputenc}
\usepackage[small]{caption}
\usepackage{booktabs} 
\usepackage{makecell}

\usepackage{amsthm}
\usepackage{graphicx}
\usepackage[english]{babel}
\usepackage{subfigure}
\usepackage{caption}
\usepackage{algorithm} 
\usepackage{algorithmic} 
\usepackage{multirow} 
\usepackage{amsfonts}
\usepackage{mathrsfs}

\begin{document}
\title{HyperST-Net: Hypernetworks for Spatio-Temporal Forecasting}
\author{
	Zheyi Pan$^1$\ Yuxuan Liang$^2$ \ Junbo Zhang$^{3,4}$ \ Xiuwen Yi$^{3,4}$ \ Yong Yu$^1$ \ Yu Zheng$^{3,4}$ \\
	$^1$Shanghai Jiaotong University \\
	$^2$Xidian University \\
	$^3$JD Urban Computing Business Unit \\
	$^4$JD Intelligent City Research \\
	zhpan@apex.sjtu.edu.cn \ \{yuxliang, msjunbozhang, xiuyi\}@outlook.com\\
	yyu@apex.sjtu.edu.cn \ msyuzheng@outlook.com
}

\maketitle

\begin{abstract}
	Spatio-temporal (ST) data, which represent multiple time series data corresponding to different spatial locations, are ubiquitous in real-world dynamic systems, such as air quality readings.
	Forecasting over ST data is of great importance but challenging as it is affected by many complex factors, including spatial characteristics, temporal characteristics and the intrinsic causality between them. 
	In this paper, we propose a general framework (HyperST-Net) based on hypernetworks for deep ST models. More specifically, it consists of three major modules: a spatial module, a temporal module and a deduction module. Among them, the deduction module derives the parameter weights of the temporal module from the spatial  characteristics, which are extracted by the spatial module.
	Then, we design a general form of HyperST layer as well as different forms for several basic layers in neural networks, including the dense layer (HyperST-Dense) and the convolutional layer (HyperST-Conv).
	Experiments on three types of real-world tasks demonstrate that the predictive models integrated with our framework achieve significant improvements, and outperform the state-of-the-art baselines as well.
\end{abstract}

\section{Introduction}
	With the rising demand for safety and health-care, large amounts of sensors have been deployed in different geographic locations to provide the real-time information of the surrounding environment.  These sensors generate massive and diverse spatio-temporal (ST) data with both timestamps and geo-tags. Predicting over such data plays an essential role in our daily lives, such as human flow prediction \cite{zhang2017deep}, air quality forecasting \cite{liang2018geoman} and taxi demand prediction \cite{yao2018deep}. Generally, in the research field of ST data, we use the two following information of a specific ST object, e.g., air quality readings from a sensor and traffic volume reported by a loop detector, to make predictions. 
	\begin{itemize}
		\item \textbf{Spatial attributes}. 
		It has been well studied in many researches \cite{yuan2012discovering} that the spatial attributes (e.g., locations, categories and density of nearby points of interest) can reveal spatial characteristics of the object.
		For example, a region containing numerous office buildings tends to be a business area, while a region with a series of apartments is likely to be a residential place.
		\item \textbf{Temporal information}. It can be easily seen that the temporal information, such as historical values from itself, can reveal the temporal characteristics of the object, which contributes a lot to the prediction task \cite{hamilton1994time}. One such example is that an air quality record from a sensor is closely related to its previous readings and weather conditions \cite{liang2018geoman}.
	\end{itemize}
	
\begin{figure}[htbp!]
		\centering
		\includegraphics[width=0.99\linewidth]{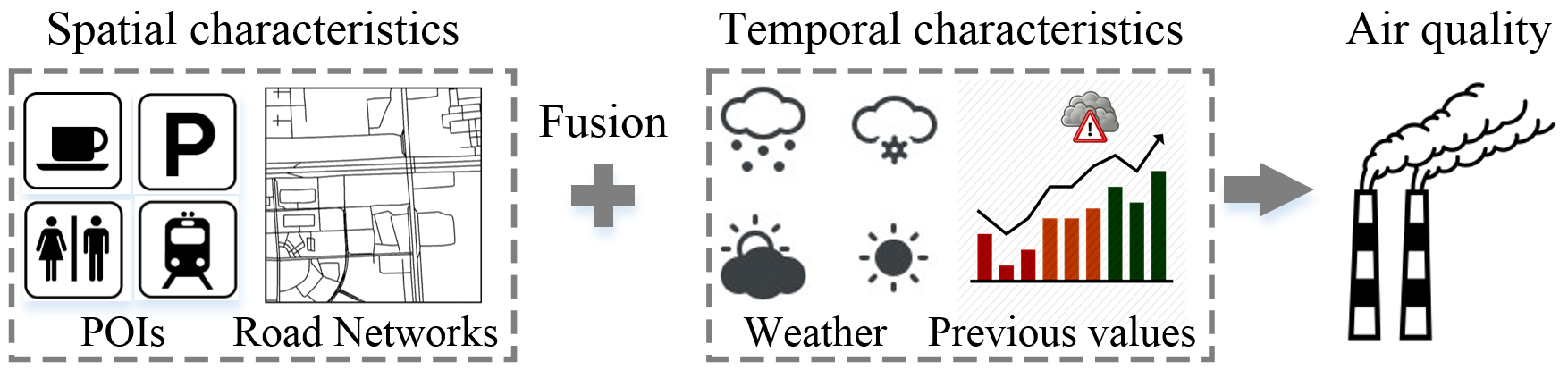}
		\caption{Example of conventional framework for ST data.}
		\label{fig:conventional:framework}
	\end{figure}

	Very recently, there has been significant growth of interests in how to combine the spatial and temporal characteristics effectively to solve real-world problems with the well-known deep learning approaches.
	Figure \ref{fig:conventional:framework} shows an example of the conventional \underline{s}patio-\underline{t}emporal \underline{net}work (ST-Net) for a typical ST application, i.e., air quality forecasts, which comprises of two modules:
	1) a spatial module to capture the spatial characteristics from spatial attributes, e.g., points of interest (POIs) and road networks;
	2) a temporal module to consider the temporal characteristics from temporal information like weather and historical readings.
	After that, the network incorporates the two kinds of characteristics by a fusion method (e.g., directly concatenation) to make predictions on future readings. However, existing fusion methods do not consider the intrinsic causality between them. 
	
\begin{figure}[htbp!]
		\centering
		\includegraphics[width=0.99\linewidth]{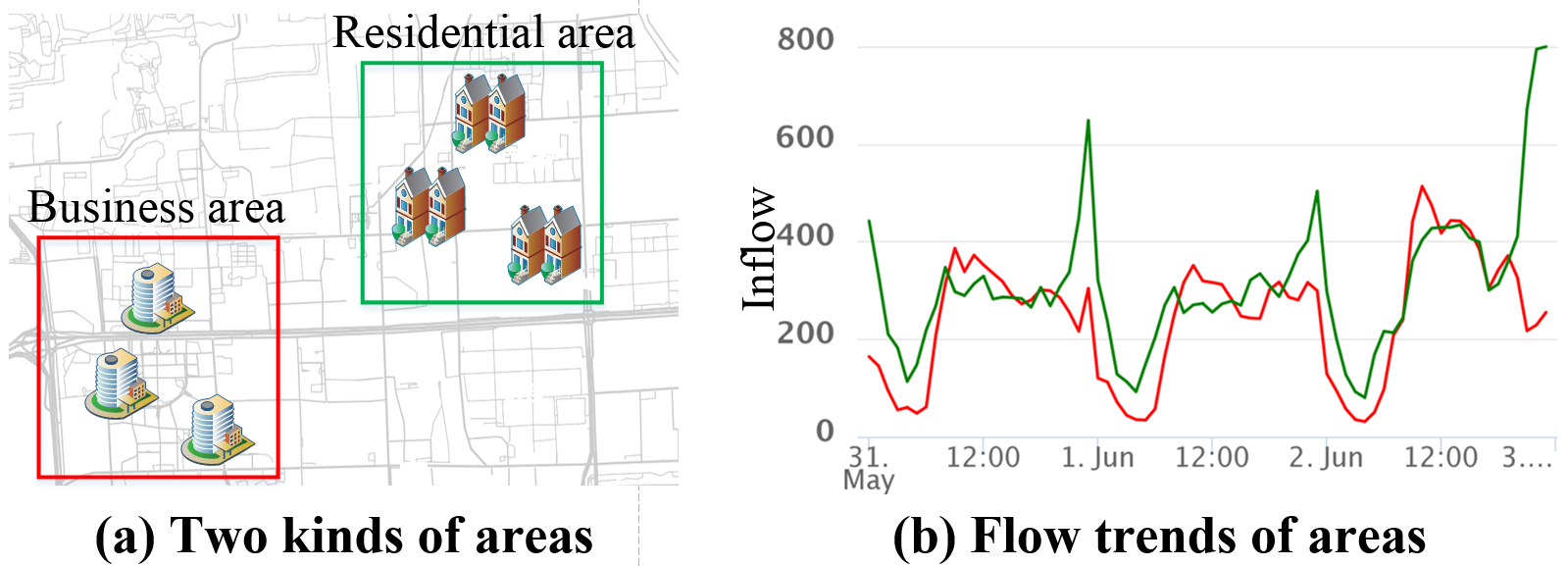}
		\caption{Taxi inflows of two typical areas, i.e., a business area and a residential area (best view in color).}
		\label{fig:flow:example}
	\end{figure}

	Generally, the \emph{intrinsic causality} between spatial and temporal characteristics of such data is crucial because it is a key towards ST forecasting.
	As depicted in Figure \ref{fig:flow:example}(a), the area in red contains numerous office buildings and expressways, which represents a business area.
	Besides, the green area with many apartments and tracks denotes a region for residents.
	In general, citizens usually commute from home to their workplaces during a day, which leads to an upward trend of inflow into the business area in the morning, but a drop-off at night.
	in contrast, Figure \ref{fig:flow:example}(b) shows that the residential area often meets a rush hour in the supper time. 
	From this example, it can be easily seen that spatial attributes (POIs and road networks) reflect the spatial characteristics, i.e. the function of a region, and further have great influence on temporal characteristics (inflow trend).
	Finally, such temporal characteristics and other time-varying information (e.g., previous values, weather) determine the future human flows simultaneously.
	Therefore, it is a non-trivial problem to capture the intrinsic causality between the spatial and temporal characteristics of objects.

\begin{figure}[htbp!]
		\centering
		\includegraphics[width=0.99\linewidth]{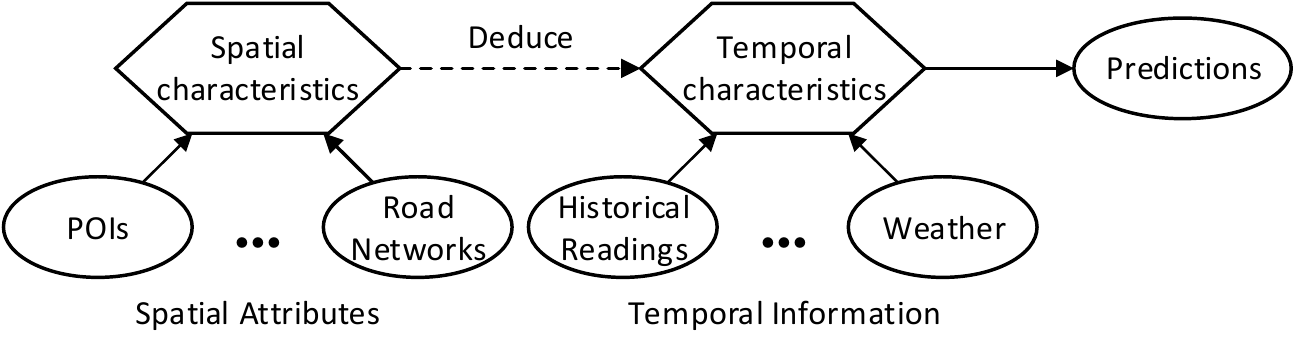}
		\caption{Insight of the proposed framework.}
		\label{fig:insight}
	\end{figure}

	Inspired by this observation, we use Figure \ref{fig:insight} to depict the interrelationship of spatial and temporal characteristics for ST forecasting.
	Firstly, spatial characteristics of an ST object are determined by its spatial attributes, like POIs and road networks.
	Secondly, considering the above causality, we deduce temporal characteristics from spatial characteristics, as shown in Figure \ref{fig:insight}.
	Finally, the predicted values are obtained from the temporal characteristics, based on temporal information (e.g., previous readings and weather). 
	
	In this paper, we propose a novel framework (HyperST-Net) based on hypernetworks to forecast ST data. The	contributions of our study are three-fold:
	\begin{itemize}
		\item We propose a novel deep learning framework, which consists of a spatial module, a temporal module, and a deduction module. 
		Specifically, the deduction module derives the parameter weights of the temporal module from the spatial characteristics, which are extracted by the spatial module. To the best of our knowledge, it is the first deep framework considering the intrinsic causality between spatial and temporal characteristics.

		\item We design a general form of HyperST layer. To make the framework more scalable and memory efficient, we further design different HyperST forms for several basic layers in neural networks, including the dense layer and the convolutional layer.
		
		\item We evaluate our framework on three representative real-world tasks: air quality prediction, traffic prediction, and flow prediction. Applying the framework for simple models (e.g., LSTM \cite{hochreiter1997long}) can significantly improve the performance, and achieve near the results of the complex hand-crafted models designed for specific tasks. Extensive experiments show that our framework outperforms the state-of-the-art baselines.
	\end{itemize}

\section{Preliminary}
	In this section, we define the notations and the studied problems, and briefly introduce the theory of Hypernetworks.
	\subsection{Notations}
		\noindent \textbf{\textit{Definition 1.}} \textit{Spatial attributes}. Suppose there are $N$ objects reporting ST data with $M$ successive time slots. We employ $\mathbf{S}=(\mathbf{s}_1,...,\mathbf{s}_N) \in \mathbb{R}^{M \times D_s}$ to represent spatial attributes (e.g., a combination of POIs and road networks features) of all objects, where $\mathbf{s}_i$ belongs to object $i$. 
		
		\noindent \textbf{\textit{Definition 2.}} \textit{Temporal information}.
		$\mathcal{T}=(\mathbf{T}_1,...,\mathbf{T}_N)$.
		We use $\mathbf{T}_i \in \mathbb{R}^{M \times D_T}$ to denote the temporal information (e.g., historical readings and weather) of the $i$-th object in a period of $M$ timestamps, where each row is a $D_T$-dimensional vector. We combine them into a tensor $\mathcal{T}=(\mathbf{T}_1,...,\mathbf{T}_N)$ to describe the temporal information of all objects.
		
		\noindent \textbf{\textit{Definition 3.}} \textit{Predicted values}. $\mathcal{L}=(\mathbf{L}_1,...,\mathbf{L}_N)$ is a tensor of labels of prediction tasks (e.g., human flows, PM2.5). $\mathbf{L}_i \in \mathbb{R} ^ {M \times D_L}$ where row $j$ is a $D_L$ dimensional vector, which indicates the readings of object $i$ at time slot $j$. 
		
		\noindent \textbf{\textit{Problem Statement.}}
		Given spatial attributes $\mathbf{S}$ and temporal information $\mathbf{T}$, we aim to find a model $f$ with parameters $\theta$, such that $f(\mathbf{S}, \mathcal{T};\theta) \rightarrow \mathcal{L}$. For simplicity, in this paper we only consider modeling $f$, such that $f(s_i, \mathbf{T}_i;\theta) \rightarrow \mathbf{L}_i$, $\forall i$.
	
	\subsection{Hypernetworks}
		Hypernetworks aim to generate the weights of a certain network from another network \cite{stanley2009hypercube}. \citeauthor{ha2016hypernetworks} explored the usage of hypernetworks for CNNs and RNNs, which can be considered as a relaxed-form of weight-sharing across multiple layers. 
		In our study, we apply this kind of framework to model the causality between spatial and temporal characteristics.

\section{Framework}
\begin{figure}[htbp!]
		\centering
		\includegraphics[width=0.99\linewidth]{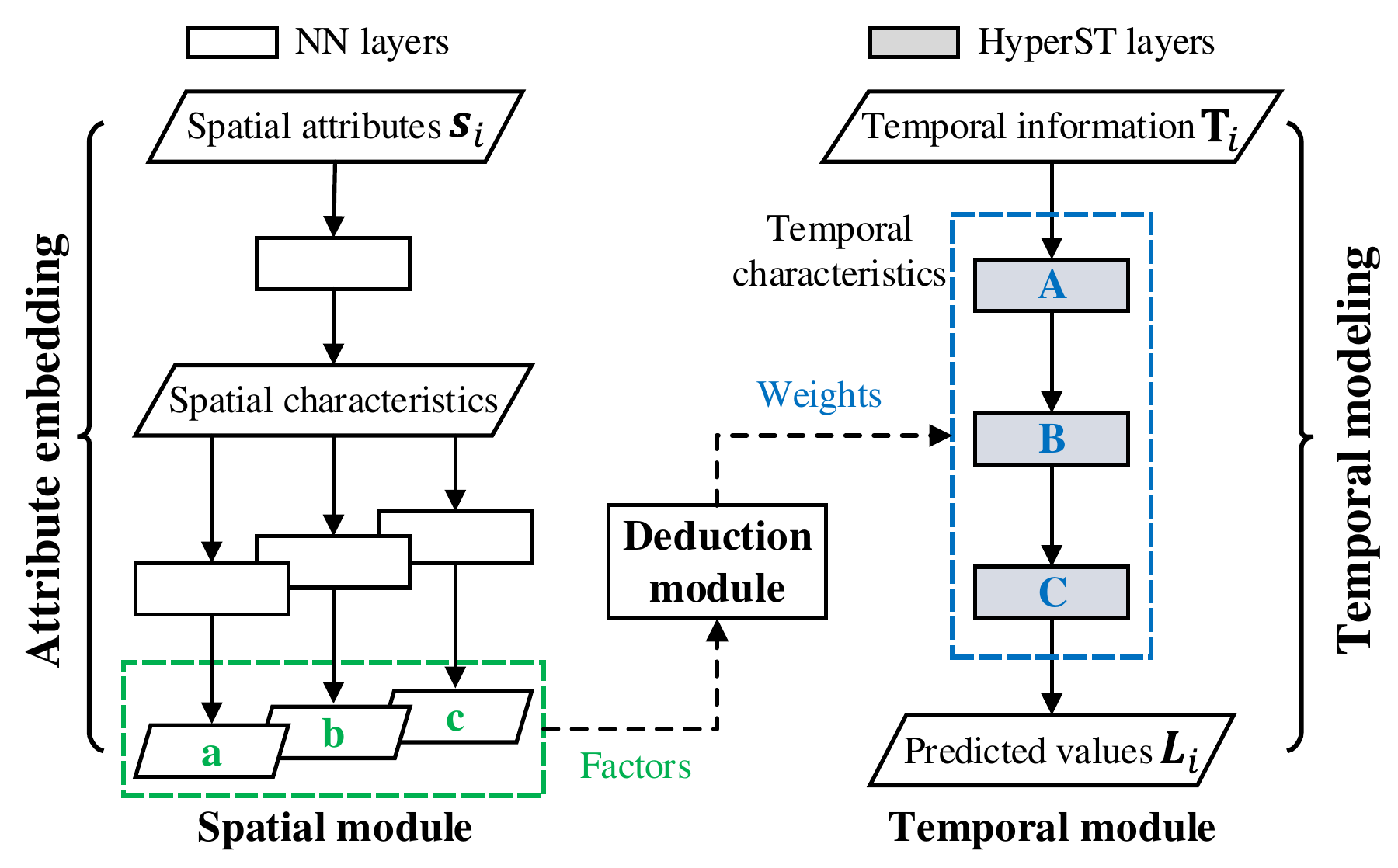}
		\caption{Framework of HyperST-Net.}
		\label{fig:framework}
	\end{figure}

	HyperST-Net consists of a spatial module, a temporal module, and a deduction module, as shown in Figure \ref{fig:framework}.
	The spatial module is used to extract spatial characteristics from spatial attributes. Once we obtain the spatial characteristics, temporal characteristics can be derived by the deduction module, i.e., we get the temporal module for time series prediction. The insights behind this framework are to capture the intrinsic causality between the spatial and temporal characteristics, so as to improve the predictive performance. We detail the three modules as follows.
	
	\noindent \textbf{Spatial module} is a two-stage module.
	As shown in Figure \ref{fig:framework}, in the first stage, spatial attributes are embedded into a low dimensional vector, i.e. spatial characteristics.
	In the second stage, it generates a series of factors (parallelograms in the green rectangle) independently, and then use them to model parameter weights of the corresponding  \underline{n}eural \underline{n}etwork (NN) layer in the temporal module by the deduction module, i.e. a $\rightarrow$ A, b $\rightarrow$ B, and c $\rightarrow$ C. 
	Therefore, the spatial module performs like a hypernetwork, allowing the spatial characteristics to play an important role in making predictions.
	
	\noindent \textbf{Temporal module} employs different kinds of HyperST layers. Compared with common NN layers (e.g., dense layer and convolution layer), the parameter weights of HyperST layers are computed by the deduction module.
	Such weights can be regarded as the temporal characteristics of the objects, which determine the future values based on temporal information.
	
	\noindent \textbf{Deduction module} bridges the spatial and temporal modules by applying a deduction process for the parameter weights, such that the intrinsic causality between spatial and temporal characteristics is well considered.
	
	In summary, the spatial attributes of various objects result in the temporal modules with different parameter weights, so as to model the distinctiveness of different objects.
	Differing from the conventional framework (ST-Net), which uses a single model for all objects, HyperST-Net is equivalent to automatically design $N$ temporal models for corresponding objects by their spatial attributes. In addition, for those objects with similar spatial attributes, the framework can deduce similar parameter weights for the temporal module. Hence, it can be seen as a relaxed-form of weight sharing across objects.
	
%
	
\section{Methodologies}
	The proposed framework consists of several HyperST layers. In this section, we first demonstrate the general form of HyperST layer, and then design specific HyperST forms of the basic layers in neural networks. 
		
	\subsection{General HyperST Layer}
\begin{figure}[htbp!]
			\centering
			\includegraphics[width=0.80\linewidth]{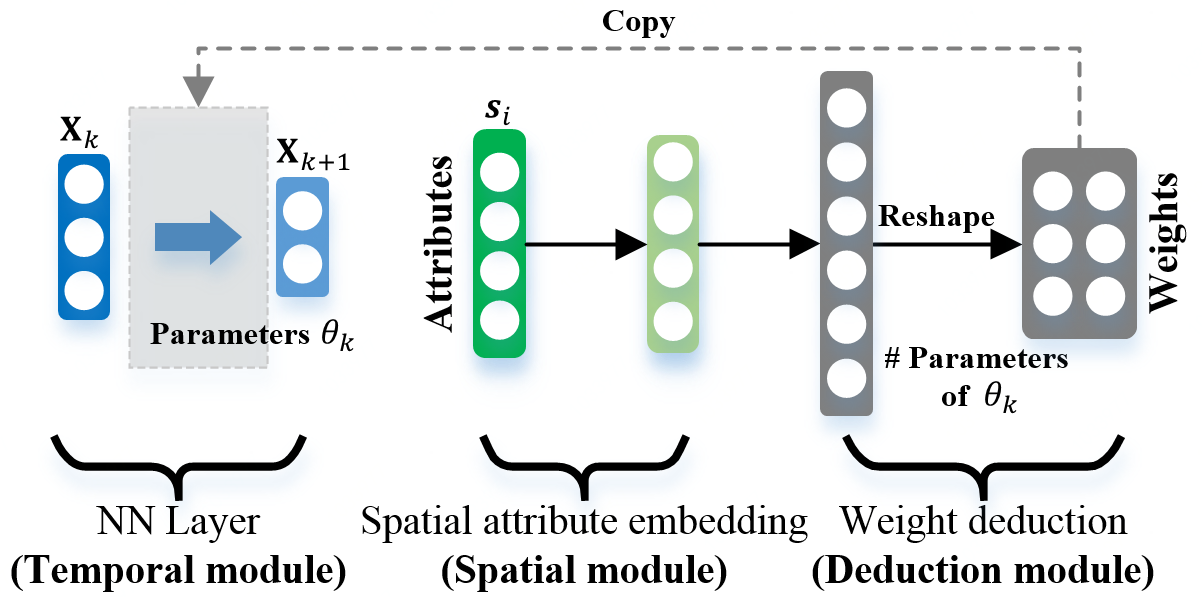}
			\caption{Illustration of a general HyperST layer}
			\label{fig:HyperST:General}
		\end{figure}

		An implementation of a general HyperST layer is shown in Figure \ref{fig:HyperST:General}. the $k$-th layer in the temporal network, denoted as $f_k$, maps the input $\mathbf{X}_{k}$ to $\mathbf{X}_{k+1}$ by a set of parameters $\theta_k$:
		\begin{equation}
			\mathbf{X}_{k+1}=f_k(\mathbf{X}_{k};\theta_k).
		\end{equation}
		$\theta_k$ can be modeled by the spatial network $g_k$ by using a set of parameters $\omega_k$:
		\begin{equation}
			\theta_k=g_k(\mathbf{s}_i; \omega_k).
			\label{eq:2}
		\end{equation}
		Figure \ref{fig:HyperST:General} depicts the spatial module first embeds spatial attributes into a vector with the dimension equals to the number of parameters in $\theta_k$.
		Then, the deduction module reshapes the vector to a tensor with the shape specified by $\theta_k$.
		Finally, the tensor, i.e., the output of HyperST layer, is used as the parameters of the NN layer in the temporal module.

		In Equation \ref{eq:2}, $\omega_k$ can be trained by back-propagation \cite{rumelhart1986learning} and gradient descent.
		Assume that the global loss function is $l(f(\mathbf{s}_i, \mathbf{T}_i;\theta),  \mathbf{L}_i)$, for example, the loss function of square error can be expressed as:
		$l(f(\mathbf{s}_i, \mathbf{T}_i;\theta), \mathbf{L}_i)=\frac{1}{2}||f(\mathbf{s}_i, \mathbf{T}_i;\theta) - \mathbf{L}_i||^2$. Then the gradient of $\omega_k$ can be expressed as:
		\begin{equation}
			\frac{\partial l}{\partial \omega_k} =
			\frac{\partial l}{\partial \mathbf{X}_{k+1}} \frac{\partial \mathbf{X}_{k+1}}{\partial \theta_k} \frac{\partial \theta_k}{\partial \omega_k} =
			\frac{\partial l}{\partial \mathbf{X}_{k+1}} \frac{\partial f_k}{\partial \theta_k} \frac{\partial g_k}{\partial \omega_k}
		\end{equation}
		
		However, the general HyperST layer introduces more parameters in practice, leading to memory usage problems. In Figure \ref{fig:HyperST:General}, suppose the number of the parameters in the original NN layer is $N_{\theta_k}$ and we use a dense layer to generate the parameters from a $d$-dimensional hidden vector. Then the total amount of introducing parameters in $\omega_k$ would be $d N_{\theta_k}$. To make the framework more scalable and memory efficient, we design HyperST forms for the dense layer and the convolutional layer in the following subsections.
		
	\subsection{HyperST-Dense}
\begin{figure}[htbp!]
			\centering
			\includegraphics[height=1.2in]{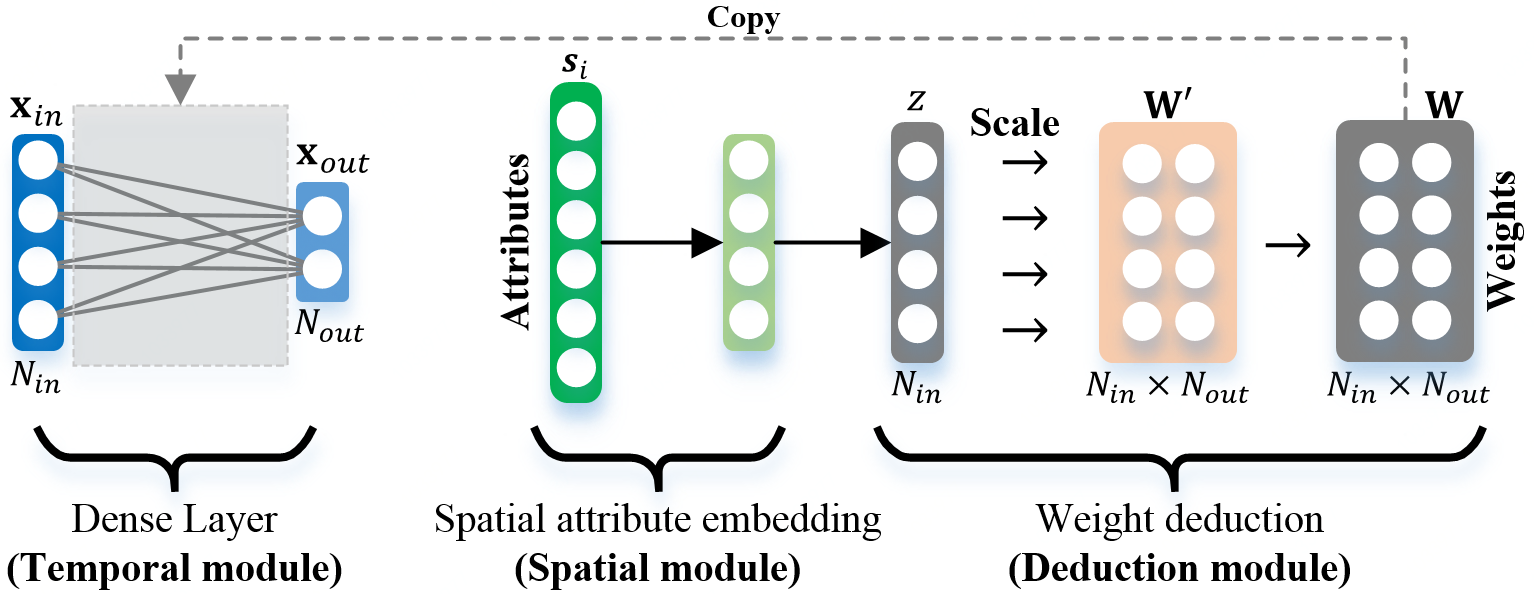}
			\caption{Illustration of HyperST-Dense}
			\label{fig:HyperST:Dense}
		\end{figure}

		As shown in Figure \ref{fig:HyperST:Dense}, the input of the dense layer in the temporal module is $\mathbf{x}_{in} \in \mathbb{R}^{N_{in}}$ and the corresponding output is $\mathbf{x}_{out} \in \mathbb{R}^{N_{out}}$. $\mathbf{x}_{out}=\mathbf{W}^{\top} \mathbf{x}_{in}$, where $\mathbf{W} \in \mathbb{R}^{N_{in} \times N_{out}}$.
		Hence, the number of parameters in $\mathbf{W}$ is $N_{in}  N_{out}$.
	
		Here, we employ the spatial module to generate a weight scaling vector $\mathbf{z} \in \mathbb{R}^{N_{in}}$, and the deduction module to scale rows of a weight matrix $\mathbf{W}'$, such that $\mathbf{W} = \mathrm{diag}(\mathbf{z}) \mathbf{W}'$, where $\mathrm{diag}(\mathbf{z})$ constructs a diagonal matrix from $\mathbf{z}$, while $\mathbf{W}'$ is learnable. If we use a dense layer to get all parameter weights from a $d$-dimensional hidden vector, the HyperST-Dense layer would contain $dN_{in}+N_{in}N_{out}$ parameters. 
		Compared with the general HyperST layer in Figure \ref{fig:HyperST:General} that introduces $dN_{in}N_{out}$ parameters, the number of parameters can be easily controlled if $d \ll N_{in}$ and $d \ll N_{out}$.
		
		For temporal model, \underline{R}ecurrent \underline{N}eural \underline{N}etwork (RNN) is widely used in practice. We can also extend the dense layer in the RNN cell to the HyperST-Dense layer. For example, the LSTM with HyperST-Dense layers (\textbf{HyperST-LSTM-D}) can be formulated as: 
		\begin{eqnarray*}
			\mathbf{f}_t & = & \sigma_g(\mathbf{W}_f^{\top} \mathrm{diag}(\mathbf{z}_0) \mathbf{x}_t + \mathbf{U}_f^{\top} \mathrm{diag}(\mathbf{z}_1) \mathbf{h}_{t-1} + \mathbf{b}_f), \\
			\mathbf{i}_t & = & \sigma_g(\mathbf{W}_i^{\top} \mathrm{diag}(\mathbf{z}_2) \mathbf{x}_t + \mathbf{U}_i^{\top} \mathrm{diag}(\mathbf{z}_3) \mathbf{h}_{t-1} + \mathbf{b}_i), \\
			\mathbf{o}_t & = & \sigma_g(\mathbf{W}_o^{\top} \mathrm{diag}(\mathbf{z}_4) \mathbf{x}_t + \mathbf{U}_o^{\top} \mathrm{diag}(\mathbf{z}_5) \mathbf{h}_{t-1} + \mathbf{b}_o), \\
			\mathbf{c}'_t & = & \sigma_c( \mathbf{W}_c^{\top} \mathrm{diag}(\mathbf{z}_6) \mathbf{x}_t + \mathbf{U}_c^{\top} \mathrm{diag}(\mathbf{z}_7) \mathbf{h}_{t-1} + \mathbf{b}_c), \\
			\mathbf{c}_t & = & \mathbf{f}_t \circ \mathbf{c}_{t-1} + \mathbf{i}_t \circ \mathbf{c}'_t, \\
			\mathbf{h}_t & = & \mathbf{o}_t \circ \sigma_c(\mathbf{c}_t),
		\end{eqnarray*}
		where $\circ$ is the element-wise multiplication, $\sigma_g$ is sigmoid function and $\sigma_c$ is hyperbolic tangent function.
		$\mathbf{z}_\eta$, as well as $\mathbf{b}_\Omega$ are vectors generated by the spatial module, while $\mathbf{W}_\Omega$ and $\mathbf{U}_\Omega$ are learnable matrices, where $\eta \in \{0,1,2,3,4,5,6,7\}$ and $\Omega \in {\{f,i,o,c\}}$.
		
	\subsection{HyperST-Conv}
\begin{figure}[htbp!]
			\centering
			\includegraphics[height=1.2in]{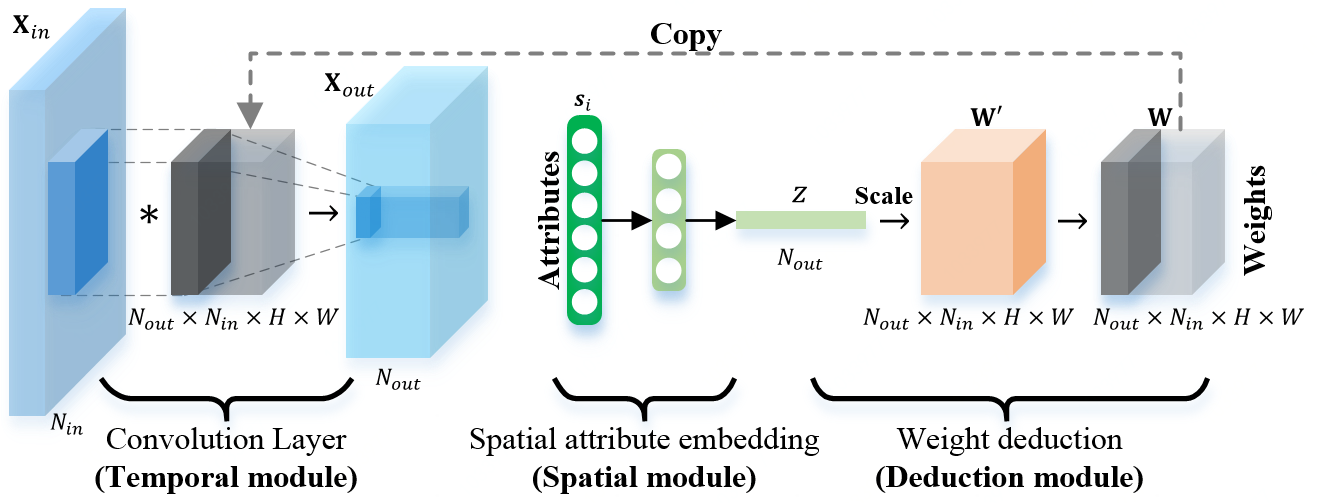}
			\caption{Illustration of HyperST-Conv}
			\label{fig:HyperST:Conv}
		\end{figure}

		As illustrated in Figure \ref{fig:HyperST:Conv}, the input of the convolution operator is a tensor $\mathbf{X}_{in}$ with $N_{in}$ channels, while the output is $\mathbf{X}_{out}$ with $N_{out}$ channels.  
		The convolution kernel is a tensor with shape of $N_{out} \times N_{in} \times H \times W$, where $H,W$ stand for the height and the width of the kernel, respectively. 
		
		To avoid introducing too many parameters, we utilize the similar method as the HyperST-Dense, to scale the weight tensor $\mathbf{W}'$. 
		In Figure \ref{fig:HyperST:Conv}, the spatial module first embeds the spatial attributes into a weight vector $\mathbf{z} \in \mathbb{R}^{N_{out}}$.
		Then in the deduction module, the kernel of the convolution operator can be expressed as $\mathbf{W}=\mathrm{diag}(\mathbf{z}) \cdot \mathbf{W}'$, where $\cdot$ is the sum of the element-wise production over the last axis of the first input and the first axis of the second input (similar to dot production of matrix), and $\mathbf{W}'$ is a learnable tensor with the shape of $N_{out} \times N_{in} \times H \times W$.
		Thus, the convolution operation can be expressed as:
		\begin{equation}
			\mathbf{X}_{out} = (\mathrm{diag}(\mathbf{z}) \cdot \mathbf{W}') * \mathbf{X}_{in},
			\label{eq:HyperST:Conv}
		\end{equation}
		where $*$ is the convolution operator. If we use a dense layer to obtain $\mathbf{z}$ from $d$-dimensional hidden vector, the number of parameters in HyperST-Conv is $dN_{out} + N_{out}N_{in}HW$. Likewise, the number of introducing parameters can be limited by modulating $d$.
		
		In general, we use the same convolutional kernel to extract features along the axes of the tensor. 
		Recall that in the field of ST forecasting, pixels (grids) in $\mathbf{X}_{in}$ indicate locations or regions with different spatial characteristics (e.g., land function). Accordingly, we propose a \textbf{location-based HyperST-Conv} layer to cope with such a scenario. Suppose $\mathbf{X}_{in}^{<i,j>}$ is a slice of tensor $\mathbf{X}_{in}$ with height $H$, width $W$ and centered at grid $(i,j)$, $\mathbf{z}^{i,j}$ is the generated weight vector of this grid, and the output vector of this grid $\mathbf{X}_{out}^{i,j}$ can be calculated as:
		\begin{eqnarray*}
			\mathbf{X}_{out}^{i,j} & = & (\mathrm{diag}(\mathbf{z}^{i,j}) \cdot \mathbf{W}') * \mathbf{X}_{in}^{<i,j>}, \\
								   & = & \mathrm{diag}(\mathbf{z}^{i,j})  \cdot (\mathbf{W}' * \mathbf{X}_{in}^{<i,j>}).
			\label{eq:location:HyperST:Conv}
		\end{eqnarray*}
		Since $\mathbf{W}'$ is shared among all grids, the location-based HyperST-Conv is equivalent to applying a conventional convolution operator with $N_{out}$ channels. It is followed by a channel-wise scaling layer, whose scaling weights $\mathbf{z}$ are generated by the spatial attributes of each grid.
		
		Moreover, location-based HyperST-Conv can be applied for other typical types of convolution operators such as graph convolution \cite{defferrard2016convolutional} and diffusion convolution \cite{li2018diffusion}.

\section{Evaluation} \label{sec:experiments}
	\subsection{Experimental Settings}	
		\subsubsection{Dataset}
		In this paper, we evaluate HyperST-Net on three representative spatio-temporal tasks as follows:
		\begin{itemize}
			\item \textbf{Air quality prediction} \cite{liang2018geoman}: 
			The air quality dataset is composed of massive readings of different pollutants (e.g., PM2.5, SO2), as well as meteorological records. We extract the density of POIs around a sensor as its spatial attributes. Based on the previous air quality readings, POI features and weather conditions, we make predictions on PM2.5 in the next 6 hours. The dataset is partitioned along the time axis into non-overlapped training, validation and test set by the ratio of 8:1:1.
			\item \textbf{Traffic prediction} \cite{li2018diffusion}:
			The traffic dataset METR-LA \cite{jagadish2014big} contains 207 sensors with their readings collected from March 1st, 2012 to June 30th, 2012. Along the timeline, we partition such dataset into non-overlapped training, validation and test data by the ratio of 7:1:2. Moreover, for each sensor, we use its GPS coordinates and the road network distance from its k-nearest neighbors ($k=4$) to itself as the spatial attributes.
			\item \textbf{Flow prediction:}
			Collected from taxicabs that travel around the city, TaxiBJ dataset \cite{yuan2010t} consists of tremendous amounts of trajectories from Feb. 1st, 2015 to Jun. 2nd 2015. We first splits the Beijing city (the lower-left GCJ-02 coordinates: $39.83,116.25$, the upper-right: $40.12,116.64$) into $32 \times 32$ individual regions, and then count the hourly inflow and outflow of regions \cite{zhang2017deep}. Considering the historical inflow and outflow together with the features of POIs and road networks, we make the short-term prediction on inflow and outflow in the next timestamp. Likewise, we follow the partition rule in the first dataset to obtain training, validation and test data.
		\end{itemize}
			
		\subsubsection{Metrics}
		We use two criteria to evaluate the framework performance in the three tasks: the rooted mean squared error (RMSE) and the mean absolute error (MAE).
		
		\subsubsection{Variants}
		To verify the effectiveness of our framework, we implement five variants of it as follows:
		\begin{itemize}
			\item \textbf{HyperST-LSTM-G}. The general version of LSTM with HyperST, which adopts spatial modules to generate all the parameter weights of dense layers in LSTM cell.
			\item \textbf{HyperST-LSTM-D}. It simply replaces the dense layers in the standard LSTM by HyperST-Dense layers.
			\item \textbf{HyperST-CNN}. In this variant, we use stacked location-based HyperST-Conv layers to build a new network for grid-based flow prediction. 
			\item \textbf{HyperST-GCGRU}.
			Since the objects in traffic prediction are interconnected with each other in the format of road networks, a graph convolution method is utilized to capture the geographical correlation between them.
			Similar to the location-based HyperST-Conv layer, we add channel-wise scaling layers after graph convolution operators \cite{li2018diffusion} in the GCGRU cell.
			\item \textbf{HyperST-DCGRU}. We substitute diffusion convolution operators for graph convolution operators in the former variant, denoted as HyperST-DCGRU. 
		\end{itemize}
		The details are shown in Table \ref{tab:settings}, where the notation ($n_1$-$n_2$-...) indicates the number of hidden units or channels ($n_i$ corresponds the $i$-th layer).
		\begin{table}[htbp!]
			\centering
			\setlength{\tabcolsep}{3pt}
			\scriptsize
			\caption{The detail structures of the HyperST-Nets.}
			\begin{tabular*}{0.999\linewidth}{cc|c|c|c}
				\toprule
				\textbf{Method}		&	\textbf{Module}	&	\textbf{AQ prediction}	&	\textbf{Traffic prediction}	&	\textbf{Flow prediction}	\\
				\midrule	
				
				HyperST-			&	\textbf{Spatial}	&	Dense(16-8-2)				&	-							&	Dense(64-4-16-2)			\\
				LSTM-G				&	\textbf{Temporal}	&	LSTM(16-16)					&	-							&	LSTM(32-16)					\\
				
				\hline
				
				HyperST-			&	\textbf{Spatial}	&	Dense(16-8-4)				&	Dense(32-8-4)				&	Dense(64-16-16-8)			\\
				LSTM-D				&	\textbf{Temporal}	&	LSTM(32-32)					&	LSTM(128-128)				&	LSTM(32-32)					\\
				
				\hline
				
				HyperST-			&	\textbf{Spatial}	&	-							&	Dense(64-8-8-4)				&	-							\\
				GCGRU				&	\textbf{Temporal}	&	-							&	GCGRU(64-64)				&	-							\\
				
				\hline
				HyperST-			&	\textbf{Spatial}	&	-							&	Dense(64-8-8-4)				&	-							\\
				DCGRU				&	\textbf{Temporal}	&	-							&	DCGRU(64-64)				&	-							\\
				
				\hline
				
				HyperST-			&	\textbf{Spatial}	&	-							&	-							&	Dense(64-16-8-8)			\\
				CNN					&	\textbf{Temporal}	&	-							&	-							&	Conv3x3(64-32)				\\
				
				\bottomrule
				
			\end{tabular*}
			\label{tab:settings}
		\end{table}

			\begin{table*}[htbp!]
				\centering
				\setlength{\tabcolsep}{5pt}
				\scriptsize
				\caption{The results of air quality prediction, where the baselines refer to the work \cite{liang2018geoman}.}
				\begin{tabular*}{0.95\linewidth}{c ccccc ccccc cc}
					\toprule
					\multirow{2}{*}{\textbf{Metric}}	&	\multicolumn{5}{c}{Classical methods}	& 	\multicolumn{5}{c}{Conventional Deep models}	&	\multicolumn{2}{c}{\textbf{HyperST-Nets}}	\\
					\cmidrule(lr){2-6} \cmidrule(lr){7-11} \cmidrule(lr){12-13}
					&	ARIMA	&	VAR		&	GBRT 	&	FFA		&	stMTMVL	&	stDNN	&	LSTM	&	Seq2seq	&	DA-RNN	&	GeoMAN	&	HyperST-LSTM-G	&	HyperST-LSTM-D	\\
					\midrule
					\textbf{MAE}	& 	20.58	& 	16.17	&	15.03	&	15.75	&	19.26	&	16.49	&	16.70	&	15.09	&	15.17	&	14.08	&	13.97			&	\textbf{13.92}			\\
					\textbf{RMSE}	&	31.07	&	24.60	&	24.00	&	23.83	&	29.72	&	25.64	&	24.62	&	24.55	&	24.25	&	22.86	&	23.27			&	\textbf{22.73}			\\
					\bottomrule
				\end{tabular*}
				\label{tab:air}
			\end{table*}
			
			\begin{table*}[htbp!]
				\centering
				\setlength{\tabcolsep}{5pt}
				\scriptsize
				\caption{The results of traffic prediction, where the baselines refer to the work \cite{li2018diffusion}.}
				\begin{tabular*}{0.95\linewidth}{cc cccc cccc ccc}
					\toprule
					\multirow{2}{*}{\textbf{Time}}	&	\multirow{2}{*}{\textbf{Metric}}	&	\multicolumn{4}{c}{Classical methods}	& 	\multicolumn{4}{c}{Conventional Deep models}	&	\multicolumn{3}{c}{\textbf{HyperST-Nets}}	\\
					\cmidrule(lr){3-6} \cmidrule(lr){7-10} \cmidrule(lr){11-13}
					&					&	HA		&	ARIMA	&	VAR		&	SVR		&	FNN		&	LSTM	&	GCRNN	&	DCRNN	&	HyperST-LSTM-D	&	HyperST-GCGRU	&	HyperST-DCGRU	\\
					\midrule
					\multirow{2}{*}{15 min} &	\textbf{MAE}	&	4.16	& 	3.99	&	4.42	&	3.99	&	3.99	&	3.44	&	2.80	&	2.77	&	2.84			&	2.75			&	\textbf{2.71}			\\
					&	\textbf{RMSE}	&	7.8		&	8.12	&	7.89	&	8.45	&	7.49	&	6.30	&	5.51	&	5.38	&	5.51			&	5.32			&	\textbf{5.23}			\\
					
					\midrule
					\multirow{2}{*}{30 min} &	\textbf{MAE}	&	4.16	& 	5.15	&	5.41	&	5.05	&	4.23	&	3.77	&	3.24	&	3.15	&	3.33			&	3.16			&	\textbf{3.12}			\\
					&	\textbf{RMSE}	&	7.8		&	10.45	&	9.13	&	10.87	&	8.17	&	7.23	&	6.74	&	6.45	&	6.78			&	6.44			&	\textbf{6.38}			\\
					
					\midrule
					\multirow{2}{*}{60 min} &	\textbf{MAE}	&	4.16	& 	6.9		&	6.52	&	6.72	&	4.49	&	4.37	&	3.81	&	3.60	&	3.84			&	3.62			&	\textbf{3.58}			\\
					&	\textbf{RMSE}	&	7.8		&	13.23	&	10.11	&	13.76	&	8.69	&	8.69	&	8.16	&	7.59	&	7.94			&	7.61			&	\textbf{7.56}			\\
					\bottomrule
				\end{tabular*}
				\label{tab:traffic}
			\end{table*}
			
			\begin{table*}[htbp!]
				\centering
				\setlength{\tabcolsep}{3.8pt}
				\scriptsize
				\caption{The results of flow prediction.}
				\begin{tabular*}{0.95\linewidth}{c cccc ccccc ccc}
					\toprule
					\multirow{2}{*}{\textbf{Metric}}		&	\multicolumn{4}{c}{Classical methods}	& 	\multicolumn{5}{c}{Conventional Deep models}	&	\multicolumn{3}{c}{\textbf{HyperST-Nets}}	\\
					\cmidrule(lr){2-5} \cmidrule(lr){6-10} \cmidrule(lr){11-13}
					&	HA		&	ARIMA	&	VAR 	&	SVR		&	LSTM	&	ST-LSTM	&	ST-CNN	&	ConvLSTM	&	ST-ResNet	&	HyperST-CNN	&	HyperST-LSTM-G	&	HyperST-LSTM-D		\\
					\midrule
					\textbf{MAE}		& 	26.11	&	28.18	&	25.24	&	23.65	&	16.71	&	15.97	&	15.92	&	16.18		&	15.64		&	15.64		&	15.41					&	\textbf{15.36}				\\
					\textbf{RMSE}		&	56.57	&	61.32	&	53.01	&	36.91	&	31.93	&	30.04	&	30.08	&	30.08		&	29.99		&	30.22		&	\textbf{29.59}			&	30.17						\\
					\bottomrule
				\end{tabular*}
				\label{tab:flow}
			\end{table*}

		\subsection{Baselines}
			We compare HyperST-Net with the following baselines:
			\begin{itemize}
				\item \textbf{HA}: Historical average.
				\item \textbf{ARIMA} \cite{box1970distribution}: A well-known method for time series prediction.
				\item \textbf{VAR} \cite{zivot2006vector}: Vector Auto-Regressive, which can capture pairwise relationships among objects. 
				\item \textbf{SVR} \cite{smola2004tutorial}: A version of SVM for performing nonlinear regression.
				\item \textbf{GBRT} \cite{friedman2001greedy}: Gradient Boosting Regression Tree, which is an ensemble approach for regression tasks.
				\item \textbf{FFA} \cite{zheng2015forecasting}: A multi-view based hybrid model considers spatio-temporal dependencies and sudden change simultaneously to forecast sensor’s reading.
				\item \textbf{stMTMVL} \cite{liu2016predicting,liu2016urban}: A general model for co-predicting the time series of different objects based on multi-task multi-view learning.
				\item \textbf{FNN}: Feed Forward Neural Network, which contains multiple dense layers to fit the temporal information to the observations of objects.
				\item \textbf{LSTM} \cite{gers1999learning}: A practical variant of RNN to model time series.
				\item \textbf{Seq2seq} \cite{sutskever2014sequence}: This model uses a RNN encoder to encode the temporal information, and another RNN decoder to make prediction.
				\item \textbf{stDNN} \cite{zhang2016dnn}: a deep neural network based model for ST prediction tasks.
				\item \textbf{ST-LSTM}: LSTM for ST data, which fuses spatial and temporal information by concatenating the hidden states, as shown in Figure \ref{fig:conventional:framework}.
				\item \textbf{ST-CNN}: Convolution neural network for ST data, which fuses spatial and temporal information by concatenating the feature maps along channel axis. 
				\item \textbf{ST-ResNet} \cite{zhang2017deep}: the state of art model for grid-based urban flow prediction. 
				\item \textbf{DA-RNN} \cite{qin2017dual}: a dual-staged attention model for time series prediction.
				\item \textbf{GeoMAN} \cite{liang2018geoman}: a multi-level-attention-based RNN model for ST prediction, which is the state-of-the-art in the air quality prediction task.
				\item \textbf{GCGRU}: Graph Convolutional GRU network, which uses graph convolution \cite{defferrard2016convolutional} in GRU \cite{chung2014empirical} cells, to make time series prediction on graph structures.
				\item \textbf{DCGRU} \cite{li2018diffusion}:  Diffusion Convolution GRU network, which uses diffusion convolution in GRU cells. It is the state-of-the-art in the traffic prediction.
			\end{itemize}
			Due to the distinctiveness of the three tasks, we select a subset of baselines from above list for each task respectively. We test different hyperparameters for them all, finding the best setting for each baseline.
		
		\subsection{Results}
			\subsubsection{Air Quality Prediction}
				As depicted in Table \ref{tab:air}, HyperST-LSTM-D achieves the best performance among all the methods. Compared with standard models like GBRT, LSTM and Seq2seq, this variant shows at least 7.4\% and 5.3\% improvements on MAE and RMSE respectively. In particular, it significantly outperforms the basic LSTM by 16\% on MAE. in contrast, hand-crafted models for ST forecasting (i.e. DA-RNN and GeoMAN) also work well in this task, but they still show inferiority against HyperST-LSTM-D. This fact demonstrates the advantages of our framework against basic models integrated with extran STructures (i.e., attention mechanism) for spatio-temporal modeling. Besides, the performance of HyperST-LSTM-G is very close to the best one, but generating all parameter weights of the temporal network results in heavy computational costs and lower predictive performance due to its massive parameters.
				
			\subsubsection{Traffic Prediction}
				Table \ref{tab:traffic} illustrates the experimental results in traffic prediction task. It can be easily seen that HyperST-LSTM-D achieves at least 12\% and 8\% lower MAE and RMSE than the simple models (FNN, LSTM) respectively. For the models with more complex structures, i.e., GCGRU and DCGRU, their HyperST versions show superiority as well. That is because our proposed framework enables such models to capture the intrinsic causality between the spatial and temporal characteristics.
		 
			\subsubsection{Flow Prediction}
				In Table \ref{tab:flow}, we present the results among different methods in terms of flow prediction. The deep models significantly outperform the traditional methods. Compared with their conventional ST versions, the three simple HyperST-Nets decrease the MAE of the predicted flow, i.e., they perform better than they used to be. In specific, HyperSTLSTM-D achieves the lowest MAE and shows 3.8\% improvements in prediction against ST-LSTM, while HyperST-LSTM-G achieves the best performance on RMSE. 

		\subsubsection{Overall Discussion}
\begin{figure}[htbp!]
				\centering
				\includegraphics[width=0.95\linewidth]{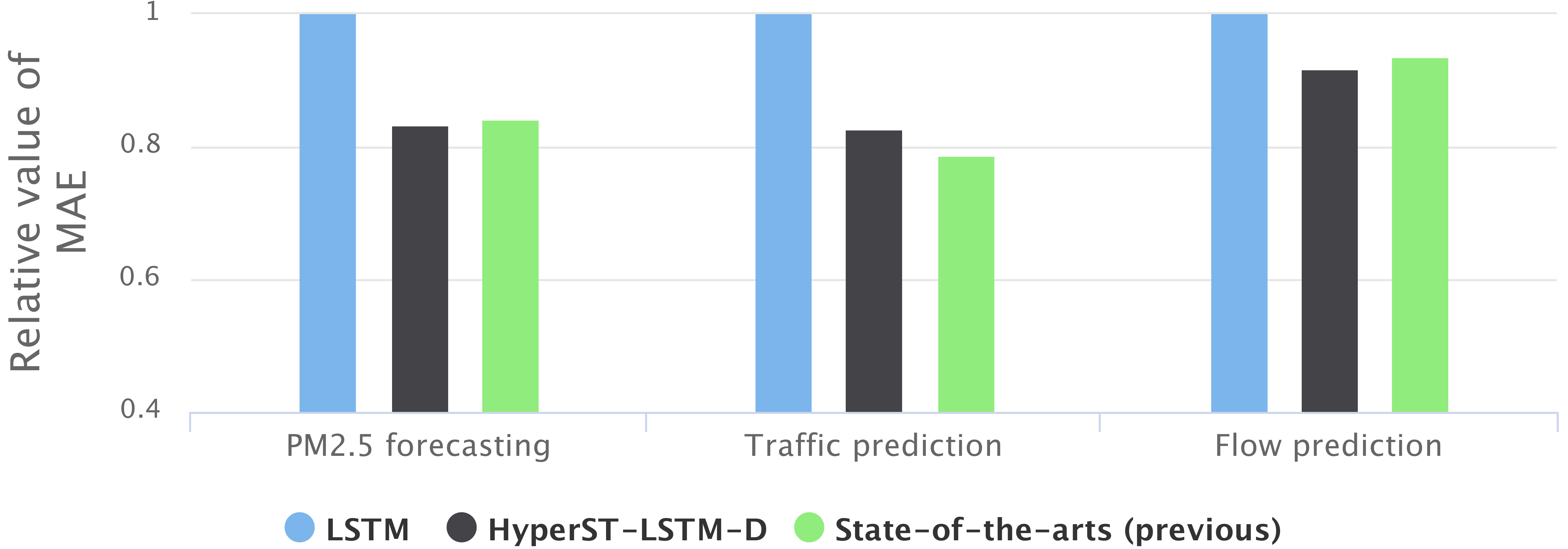}
				\caption{Improvements of simple models integrated with our framework (best view in color).}
				\label{fig:comparison}
			\end{figure}

			To investigate the effectiveness of our framework, we also compare the results of LSTM as well as the state-of-the-art methods with HyperST-LSTM-D in each task. As shown in Figure \ref{fig:comparison}, the y-axis indicates the relative value of MAE, which is computed as: the MAE of the selected model divides by that of LSTM. Primarily, for the standard LSTM, applying HyperST-Net to it brings a 16.6\%, 17.4\% and 8.1\% improvements on the three predictive tasks separately. Besides, the performance of HyperST-LSTM-D is extremely close to (even better than) the complex structured models for each task. This case demonstrates that our proposed framework significantly enhances the performance of simple models like LSTM by providing such an easy-implemented plugin for them.

	\subsection{Case Study}
		We perform a case study on learning representation for spatial attributes in the flow prediction task, to show that the HyperST-Net is capable of capturing the intrinsic causality between spatial and temporal characteristics. 
		
\begin{figure}[tbp!]
			\centering
			\includegraphics[width=0.99\linewidth]{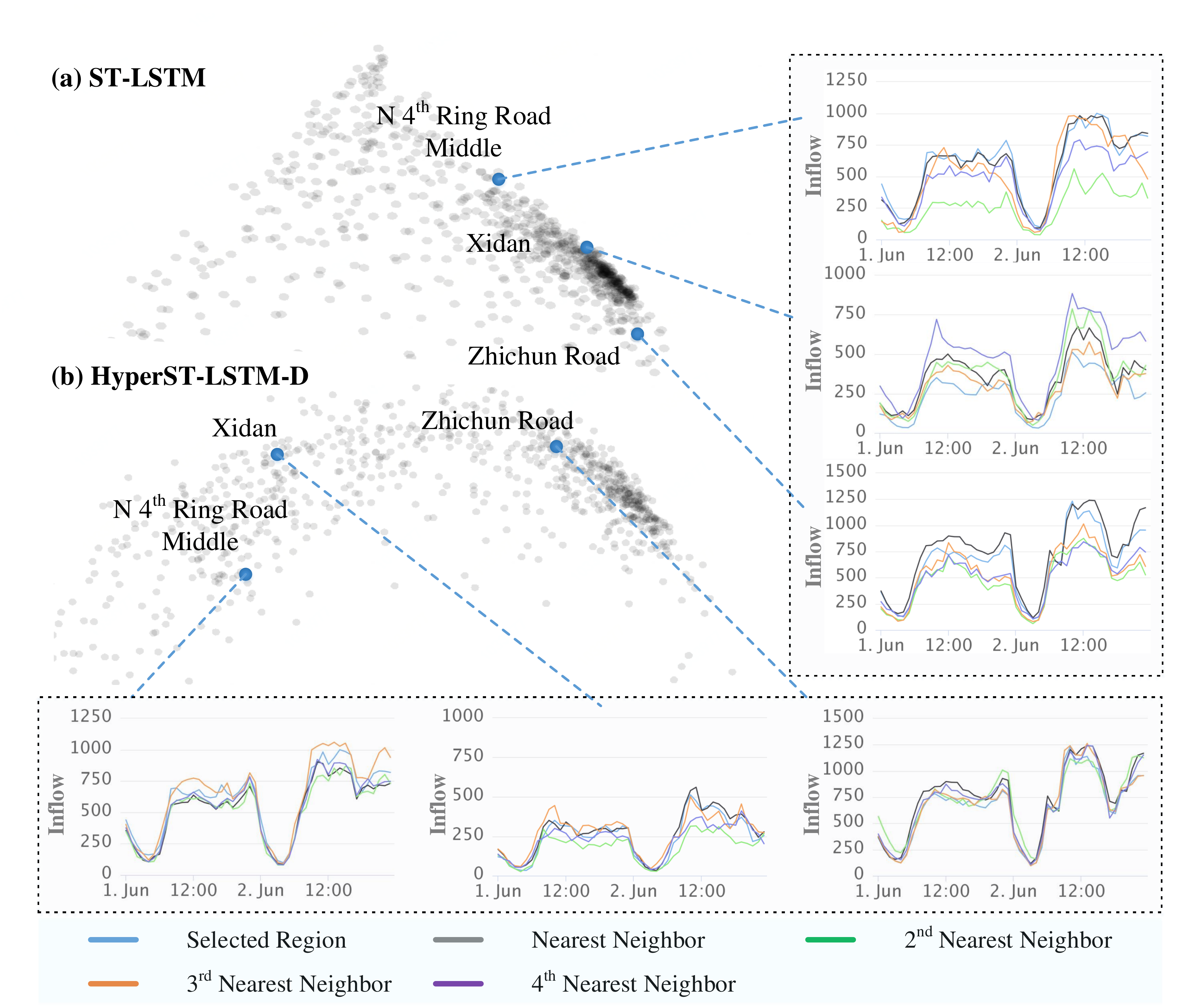}
			\caption{Embedding visualization.}
			\label{fig:case}
		\end{figure}

		As shown in Figure \ref{fig:case}, we first plot the embedding space (dimension=16) of spatial attributes for HyperST-LSTM-D and ST-LSTM, by using PCA to reduce the dimension. Most points of HyperST-LSTM-D lie in a smooth manifold, while most points of ST-LSTM are concentrated on some sharp edges, which means that the points on the same edges substantially are in extremely low dimension, and the embedding space is hard to distinguish those concentrated points which have different characteristics.
	
		Besides, we further select three representative areas with different functions as follows: 1) the region in the vicinity of Zhichun Road, which contains large amounts of apartments; 2) the region near North 4th Ring Road Middle with many expressways; 3)  the region of Xidan, which acts as a business areas full of office buildings. For each region, we first choose four nearest neighbors of itself in Euclidean space, and then plot the inflow of them in a period of two days, i.e., from June 1st, 2015 to June 2nd, 2015. 
		As shown in Figure \ref{fig:case} (a), the neighbors' flows deviate from the flow of the selected region.
		While in Figure \ref{fig:case} (b), the flow of the selected region is similar to the flow of its neighbor.
		The case strongly verifies that HyperST-Net can capture the intrinsic causality between spatial and temporal characteristics of objects. 
	
\section{Related Work}	
	\noindent \textbf{Deep Learning for ST Forecasting}.
	Deep learning technology \cite{lecun2015deep} powers many application in modern society. CNN \cite{lecun1995convolutional} is successfully used for modeling spatial correlation, especially in the field of computer vision \cite{krizhevsky2012imagenet}. RNN \cite{williams1989learning} achieves great advance in modeling sequential data, e.g. machine translation \cite{sutskever2014sequence}.

	Recently, in the field of spatio-temporal data, various work focuses on designing deep learning framework for capturing spatial correlations and temporal dependencies simultaneously.
	\citeauthor{zhang2016dnn,zhang2017deep} employ CNNs to capture spatial correlations of regions and temporal dependencies of human flows. 
	Very recent studies \cite{song2017end,zhao2017video,liang2018geoman} use attention model and RNN to capture spatial correlations and temporal dependencies, respectively. 
	\citeauthor{kong2018hst} propose to add spatio-temporal factors into the gates of RNN. 
	\citeauthor{xingjian2015convolutional,li2018diffusion} combines convolutional-based operator and RNN to model the spatio-temporal data. 
	However, the aforementioned deep learning methods for spatio-temporal data fuse the spatial information and temporal information in substance (e.g., concatenate the hidden states), without considering the intrinsic causality between them. To this end, we are the first to propose a general framework for modeling such causality, so as to improve the predictive performance.
	
	\noindent \textbf{Hypernetworks}.
	A Hypernetwork \cite{stanley2009hypercube} is a neural network used to parameterize the weights of another network (i.e., the main network), whose weights are some function (e.g. a multilayer perceptron \cite{rosenblatt1958perceptron}) of a learned embedding, such that the number of learned parameters is smaller than the full number of parameters.
	Recently, \cite{ha2016hypernetworks} explored the usage of hypernetworks for CNNs and RNNs, which can be regarded as a relaxed-form of weight-sharing across multi-layers. To the best of our knowledge, no prior work studies our problem from a hypernetwork perspective.
	
\section{Conclusion} 
	In this paper, we propose a novel framework for spatio-temporal forecasting, which is the first to consider the intrinsic causality between spatial and temporal characteristics based on deep learning. 
	Specifically, our framework consists of three major modules, i.e. a spatial module, a temporal module, and a deduction module. 
	The first module aims to extract spatial characteristics from spatial attributes. Once we obtain the spatial characteristics, temporal characteristics can be derived by the deduction module, i.e., we get the temporal module for time series prediction. 
	We design a general form of HyperST layer, which is applicable to common types of layers in neural networks.
	To reduce the complexity of networks integrated with the framework, we further design HyperST forms for the basic layers in deep learning, including dense layer, convolutional layer, etc.
	We evaluate our framework on three real-world tasks and the experiments show that the performance of simple networks (e.g. standard LSTM) can be significantly improved by integrating our framework, for example, applying it to standard LSTM brings a 16.6\%, 17.4\% and 8.1\% improvements on the above tasks separately.
	Besides, our models achieve the best predictive performance against all the baselines in terms of two metrics (MAE and RMSE) simultaneously.
	Finally, we visualize the embeddings of the spatial attributes, showing the superiority of modeling the intrinsic causality between spatial and temporal characteristics. 
	
\bibliography{reference}
\bibliographystyle{aaai}

\end{document}